\pdfoutput=1
\documentclass[11pt]{article}

\usepackage[final]{acl}

\usepackage{times}
\usepackage{latexsym}

\usepackage[T1]{fontenc}
\usepackage[utf8]{inputenc}

\usepackage{microtype}

\usepackage{inconsolata}

\usepackage[edges]{forest}
\definecolor{hidden-draw}{RGB}{205, 44, 36}
\definecolor{hidden-blue}{RGB}{194,232,247}
\definecolor{hidden-orange}{RGB}{243,202,120}
\definecolor{hidden-yellow}{RGB}{242,244,193}
\definecolor{tree-level-1}{RGB}{245,20,85}
\definecolor{tree-level-2}{RGB}{246,86,118}
\definecolor{tree-level-3}{RGB}{248,177,193}
\definecolor{tree-leaf}{RGB}{176,230,198}

\usepackage{graphicx}
\usepackage{amsmath} 
\usepackage{booktabs} 

\title{Contextual Compression in Retrieval-Augmented Generation for Large Language Models: A Survey}

\author{Sourav Verma  \\
  IBM Watsonx Client Engineering, India \\
  \texttt{sourav.verma@ibm.com} | \texttt{souravv.vermaa@gmail.com} }

\begin{document}
\maketitle
\begin{abstract}
Large Language Models (LLMs) showcase remarkable abilities, yet they struggle with limitations such as hallucinations, outdated knowledge, opacity, and inexplicable reasoning. To address these challenges, Retrieval-Augmented Generation (RAG) has proven to be a viable solution, leveraging external databases to improve the consistency and coherence of generated content, especially valuable for complex, knowledge-rich tasks, and facilitates continuous improvement by leveraging domain-specific insights. By combining the intrinsic knowledge of LLMs with the vast, dynamic repositories of external databases, RAG achieves a synergistic effect.
However, RAG is not without its limitations, including a limited context window, irrelevant information, and the high processing overhead for extensive contextual data. In this comprehensive work, we explore the evolution of Contextual Compression paradigms, providing an in-depth examination of the field. Finally, we outline the current challenges and suggest potential research and development directions, paving the way for future advancements in this area.~\footnote{Resources are available at \url{https://github.com/SrGrace/Contextual-Compression}}
\end{abstract}

\section{Introduction}
The pioneering accomplishments of large language models (LLMs) have galvanized research initiatives across both industrial and academic spheres. These LLMs showcase their capacity to converse with humans in a natural and articulate manner, excelling across various tasks such as document summarization, Q\&A systems, conversational AI, and coding assistants. Despite their advancements, LLMs continue to struggle with tasks that require specialized knowledge or domain-specific expertise. \cite{kandpal2023large}. Notably, they may produce “hallucinations” \cite{zhang2023siren} when confronted with out-of-scope queries or requests that necessitate up-to-date knowledge. To address these challenges, Retrieval-Augmented Generation (RAG) leverages external knowledge bases to retrieve relevant document snippets, utilizing semantic similarity metrics to identify the most pertinent information. By tapping into external knowledge sources, RAG successfully alleviates the issue of generating inaccurate content, thereby increasing the reliability of LLMs and paving the way for their widespread adoption in real-world applications. 

However, RAG also has its challenges. One issue is that when retrieving relevant documents, the important information may be buried in a large amount of irrelevant text, leading to inefficient and poor responses. Another challenge is that current language models have a limited input length, which causes their performance to decline when processing lengthy documents, such as academic articles, research papers, or literary works. This constraint has fueled research into developing methods to increase the input length while maintaining the model's accuracy and efficiency.

This paper aims to shed light on the latest advancements in contextual compression methods, with a focus on their application in retrieval-based systems. Our research involves a comprehensive review of methodologies, metrics, and benchmarks, which we systematically categorize into a novel taxonomy. Our taxonomy, as shown in Figure~\ref{fig:categorization_of_compression}, presents a structured and comprehensive framework for categorizing and analyzing Contextual Compression techniques for LLMs. Our investigation involves a comprehensive analysis of established techniques, such as semantic compression, in-context auto-encoder compressors, and auto-compressors, among others. Furthermore, our research highlights the ongoing challenges in this field and provides a roadmap for future investigations. We emphasize the need for collective efforts to create a sustainable and environmentally responsible future for LLMs.

\tikzstyle{my-box}=[
 rectangle,
 draw=hidden-draw,
 rounded corners,
 text opacity=1,
 minimum height=1.5em,
 minimum width=5em,
 inner sep=2pt,
 align=center,
 fill opacity=.5,
 ]
 \tikzstyle{leaf}=[my-box, minimum height=1.5em,
 fill=hidden-orange!60, text=black, align=left,font=\scriptsize,
 inner xsep=2pt,
 inner ysep=4pt,
 ]
 \begin{figure*}[t]
	\centering
	\resizebox{\textwidth}{!}{
		\begin{forest}
			forked edges,
			for tree={
				grow=east,reversed=true,anchor=base west,parent anchor=east,
				child anchor=west,base=left,font=\small,rectangle,
				draw=hidden-draw,rounded corners,align=left,minimum width=4em,
				edge+={darkgray, line width=1pt},s sep=3pt,inner xsep=2pt,
				inner ysep=3pt,ver/.style={rotate=90, child anchor=north, parent anchor=south, anchor=center},
			},
			where level=1{text width=6em,font=\scriptsize}{},
			where level=2{text width=9em,font=\scriptsize}{},
			where level=3{text width=6.6em,font=\scriptsize}{},
			[Contextual Compression for Large Language Models, ver
    			[Semantic \\Compression 
        			[
                        Context Distillation
                        [
                        Learning by distilling context~\cite{snell2022learning}{,} Gisting~\cite{mu2024learning}, leaf, text width=30em
            		]
        			]
                        [
                        Concept Distillation
                        [
                        Compressing Long Context for Enhancing RAG with AMR-based Concept Distillation~\cite{shi2024compressing}, leaf, text width=30em
            		]
        			]
        			[
                        Prompting
                        [
                        Soft Prompts
                        [
                        The Power of Scale for PEPT~\cite{lester-etal-2021-power}{,} \\OptiPrompt~\cite{zhong-etal-2021-factual}{,} Recurrentgpt~\cite{zhou2023recurrentgpt}{,} \\
                        P-Tuning~\cite{liu-etal-2022-p}, leaf, text width=21.8em    
                        ]
                        ]
                        [
                        Prompt Compression
                        [
                        Prompt compression and contrastive conditioning~\cite{wingate2022prompt}, leaf, text width=21.8em
                        ]
                        ]
                        [
                        Task-Agnostic \\Prompt Compression
                        [
                        LLMLingua~\cite{jiang-etal-2023-llmlingua}{,} LongLLMLingua~\cite{jiang-etal-2023-longllmlingua}{,} \\
                        LLMLingua-2~\cite{wu2024llmlingua2}, leaf, text width=21.8em
                	]
                        ]
        			]
                        [
                        Efficient Attention \\Operations
                        [
                        Transformer-XL~\cite{dai2019transformer}{,} Longformer~\cite{beltagy2020longformer}{,} FlashAttention~\cite{dao2022flashattention}{,} \\
                        LongLoRA~\cite{chen2023longlora}, leaf, text width=30em
            		]
        			]
                        [
                        Extrapolation and \\Interpolation
                        [
                        Exploring length generalization in LLMs~\cite{anil2022exploring}{,} Positional Interpolation(PI)~\cite{chen2023extending}{,} \\
                        YaRN~\cite{peng2023yarn}, leaf, text width=30em	
                        ]
        			]
                        [
                        Context Window \\Extension
                        [
                        Extending context window of LLMs via semantic compression~\cite{fei2023extending}, leaf, text width=30em
            		]
        			]
    			]
    			[
                    Pre-Trained \\Language Models
                    [
                    AutoCompressors
                    [
                    Adapting LMs to compress contexts~\cite{chevalier2023adapting}, leaf, text width=30em
            	]
                    ]
                    [
                    LongNET
                    [
                    LongNET: Scaling transformers to 1B tokens~\cite{ding2023longnet}, leaf, text width=30em
            	]
                    ]
                    [
                    In-Context Auto-Encoders
                    [
                    In-context autoencoder for context compression in a LLM~\cite{ge2023context}, leaf, text width=30em
            	]
                    ]
                    [
                    RECOMP
                    [
                    Retrieve-Compress-Prepend~\cite{xu2024recomp}, leaf, text width=30em
                    ]
                    ]
    			]
    			[
                    Retrievers
                    [
                    LLMChainExtractor
                    [
                    LangChain's Method~\cite{langchain}, leaf, text width=30em
            	]
                    ]
                    [
                    EmbeddingsFilter
                    [
                    LangChain's Method~\cite{langchain}, leaf, text width=30em
            	]
                    ]
                    [
                    DocumentCompressorPipeline
                    [
                    LangChain's Method~\cite{langchain}, leaf, text width=30em
            	]
                    ]
    			]
			]
		\end{forest}
	}
	\caption{Taxonomy of Contextual Compression Methods for Large Language Models.}
	\label{fig:categorization_of_compression}
\end{figure*}
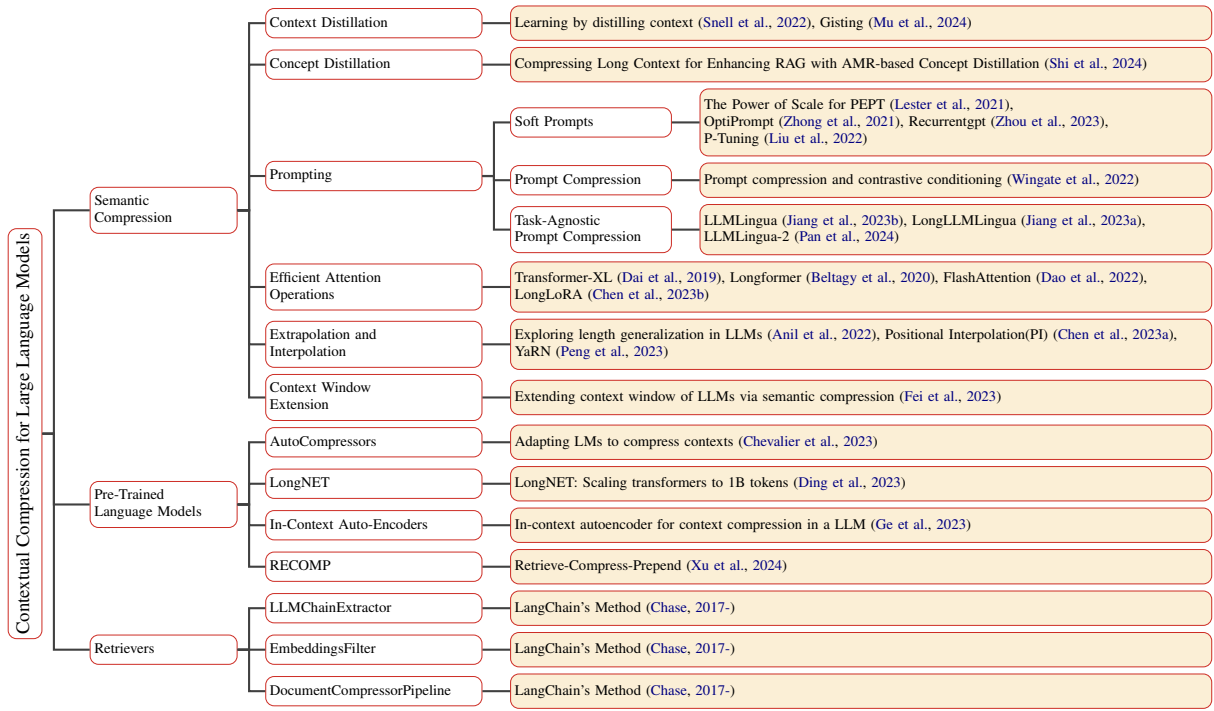

\section{Methods}
\subsection{Semantic Compression}
Semantic compression is a technique that helps identify common patterns of thought in a specific context by generalizing terms. It uses a "domain frequency dictionary" to establish the context and disambiguate multiple possible meanings of words. This approach, based on semantic networks, offers improvements over existing natural language processing techniques. 

Semantic compression reduces the number of terms in a text document by replacing less frequent terms with more general terms (their hypernyms) using a semantic network and term frequency data. This compression minimizes information loss and enables efficient processing, especially in tasks involving vector space models \cite{baeza1999modern}, \cite{erk2008structured}. It also helps address linguistic \cite{sinha2007unsupervised} challenges like polysemy and synonymy \cite{krovetz1992lexical} by replacing multiple rare terms with a single, more general concept. By using statistical analysis and frequency dictionaries, semantic compression can handle polysemic concepts more effectively and with lower error rates than other techniques.
These efforts can be summarized into five approaches: \textit{Context Distillation}, \textit{Prompting}, \textit{Efficient Attention Operations}, \textit{Extrapolation and Interpolation}, and \textit{Context Window Extension}.

\subsubsection{Context Distillation}
Recent studies have demonstrated that augmenting language models (LMs) with contextual information, such as task descriptions, illustrative examples, and explanatory notes \cite{chen2021meta}, \cite{scheurer2022learning}, can substantially enhance their performance capabilities. This approach can even facilitate zero-shot learning \cite{wei2021finetuned}, \cite{victor2022multitask} and enable models to tackle complex tasks by generating sequential reasoning steps \cite{nye2021show}, \cite{wei2022chain}, \cite{zhou2022least}.

While LMs perform better with context tokens, this advantage disappears when the tokens are removed. Additionally, processing context tokens requires extra computation, which can be a drawback. The context tokens can also be very long, and it's unclear how to handle them when they exceed the context window size. These limitations are similar to human cognitive limitations \cite{wason1974dual}, such as struggling with complex tasks and having limited working memory \cite{baddeley1992working}.

Humans overcome challenges through practice, which allows them to "distill" knowledge into habits and muscle memory. For example, learning to type a phone number becomes automatic with repetition, freeing up conscious reasoning for more complex tasks \footnote{procedural learning vs. declarative learning - \url{https://en.wikipedia.org/wiki/Procedural_knowledge}}. This process is essential for building skills and knowledge, enabling us to tackle increasingly intricate challenges.

\begin{figure}[ht]
    \centering
    \includegraphics[width=\columnwidth]{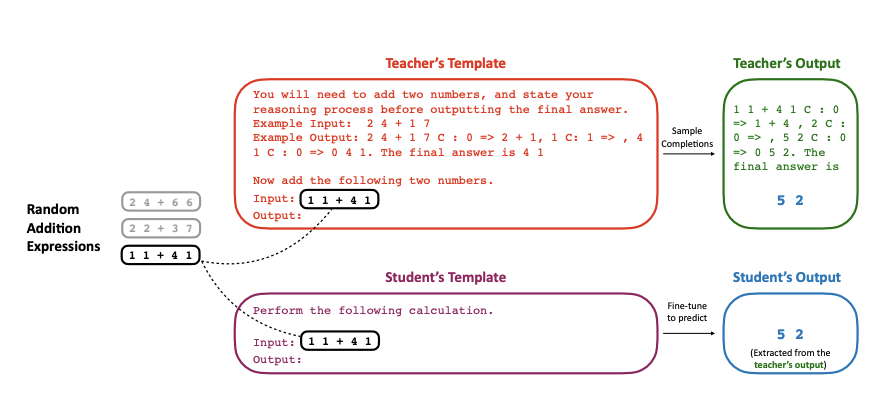}
    \caption{Internalization of step-by-step reasoning via context distillation \cite{snell2022learning}}
    \label{fig:cdistillation}
\end{figure}

\begin{figure}[ht]
    \centering
    \includegraphics[width=\columnwidth]{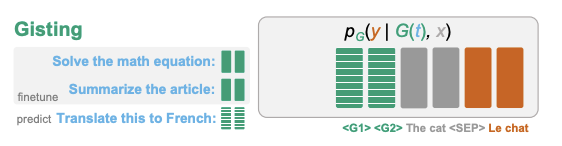}
    \caption{Gisting - Each vertical rectangle here represents a stack of Transformer activations \cite{mu2024learning}
}
    \label{fig:gisting}
\end{figure}

Researchers in NLP \cite{askell2021general}, \cite{snell2022learning} are exploring techniques to fine-tune language models, such as context distillation and "Gisting". Context distillation involves generating "practice" questions, having the model reason step-by-step, and fine-tuning it to predict answers from simpler prompts. This helps the model internalize skills, like step-by-step addition (ref Figure~\ref{fig:cdistillation}). "Gisting" \cite{mu2024learning} compresses instructions into concise key-value attention prefixes, saving computational resources and generalizing well to new tasks. As depicted in Figure~\ref{fig:gisting}, the approach involves learning a gist model by incorporating gist tokens during instruction tuning, enabling the model to handle prompt compression and instruction following simultaneously.

\subsubsection{Prompting}
\begin{figure}[ht]
    \centering
    \includegraphics[width=\columnwidth]{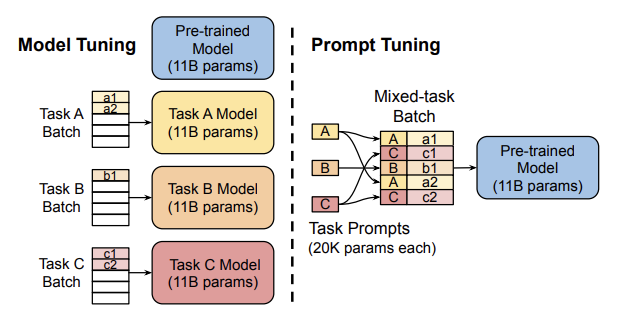}
    \caption{From 11 billion for a tuned model to just 20,480 for a tuned prompt, a reduction of over 5 orders of magnitude \cite{lester-etal-2021-power}}
    \label{fig:soft_prompting}
\end{figure}

\textbf{Soft Prompts -} As depicted in Figure~\ref{fig:soft_prompting}, soft prompt tuning enables the adaptation of pre-trained Transformers without modifying their underlying parameters, as demonstrated in recent studies \cite{lester-etal-2021-power}, \cite{zhong-etal-2021-factual}, and \cite{liu-etal-2022-p}. It entails adding novel embeddings to the input sequence and fine-tuning only these new parameters while keeping the remainder of the model's architecture frozen. This approach is categorized as a parameter-efficient fine-tuning method (PEFT) \cite{lialin2023scaling}, and bears resemblance to prefix tuning, which prepends task-specific vectors to the attention states instead of the input sequence \cite{li2021prefix}.

\noindent \textbf{Prompt Compression -} In their work, \cite{wingate2022prompt} hypothesize using a soft prompt $sp$ to compress information from a context $ctx$. They use a pre-trained LM $p_\text{LM}$ to generate continuations $cty \sim p_\text{LM}(\cdot \mid ctx)$ based on the context, and then calibrate the model's outputs with the soft prompt $sf$, $p_\text{LM}(cty \mid sf)$ to the outputs based on the context $ctx$, $p_\text{LM}(cty \mid ctx)$. They find that soft prompts effectively preserve abstract knowledge and improve guided output. Nevertheless, this method necessitates distinct optimization for each novel context, lacking the ability to leverage knowledge across analogous contexts.
\\
\textbf{Task-Agnostic Prompt Compression -}  Current methods for compressing natural language prompts remove tokens or lexical units based on information entropy from a language model like LlaMa-7B. However, using information entropy as a compression metric has two limitations: 1) it only considers unidirectional context, which may miss important information, and 2) it doesn't perfectly align with the goal of prompt compression.

\begin{figure}[ht]
    \centering
    \includegraphics[width=\columnwidth]{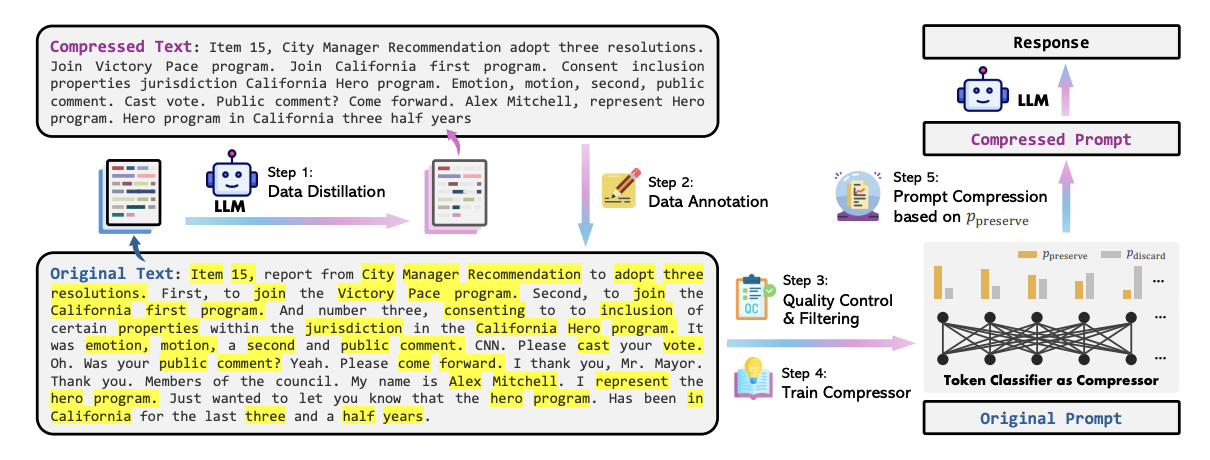}
    \caption{Overview of LLMLingua-2 \cite{wu2024llmlingua2}}
    \label{fig:llmlingua2}
\end{figure}

To address these issues, \cite{wu2024llmlingua2} propose a data distillation approach to compress prompts while retaining essential information. They introduce an extractive text compression dataset and frame prompt compression as a token classification problem (preserve or discard) (Refer to Figure~\ref{fig:llmlingua2}). The key benefits are as follows:
\begin{enumerate}
\itemsep0em
    \item \textit{Comprehensive Information Capture:} By leveraging a Transformer encoder, the method captures essential details from the full bidirectional context.
    \item \textit{Reduced Latency:} Smaller models explicitly learn the compression objective, leading to lower latency.
    \item \textit{Faithfulness:} The compressed prompt remains faithful to the original content.
\end{enumerate}

\subsubsection{Efficient Attention Operations}
The self-attention mechanism in LLMs leads to an inference cost that scales quadratically with sequence length, prompting the development of various methods to alleviate this complexity. For example:
\begin{itemize}	
\itemsep0em
    \item \textit{Transformer-XL \cite{dai2019transformer}} - employs a recurrent architecture that operates on segments, paired with a novel positional encoding technique.
    \item \textit{Longformer \cite{beltagy2020longformer}} - introduces sparse attention, scaling linearly with sequence length.
    \item \textit{FlashAttention \cite{dao2022flashattention}} - uses chunking and re-computation to avoid quadratic attention complexity.
\end{itemize}
However, these methods can be expensive to train and struggle with out-of-distribution content lengths \cite{ding2023longnet}. To address this, \textit{LongLoRA \cite{chen2023longlora}} provides a computationally efficient fine-tuning method with minimal resource requirements. For further insights, refer to the study by \cite{huang2023advancing}.

\subsubsection{Extrapolation and Interpolation}
In the field of NLP, researchers are investigating methods to extend the capabilities of existing language models, initially trained on brief texts, to process longer sequences during inference \cite{anil2022exploring}. One approach is to alter positional embeddings, which are typically designed for shorter contexts. The Rotary Position Embeddings (RoPE) from LLaMA is a key foundation for several studies in this area. For example:
\begin{itemize}	
\itemsep0em
    \item \textit{Position Interpolation (PI) \cite{chen2021meta}} applies a linear transformation to input positional indices.
    \item \textit{YaRN \cite{peng2023yarn}} leverages neural tangent kernel-inspired mechanisms to scale up the context window to 64,000 and 128,000 tokens.
\end{itemize}
\subsubsection{Context Window Extension}
\begin{figure}[ht]
    \centering
    \includegraphics[width=\columnwidth]{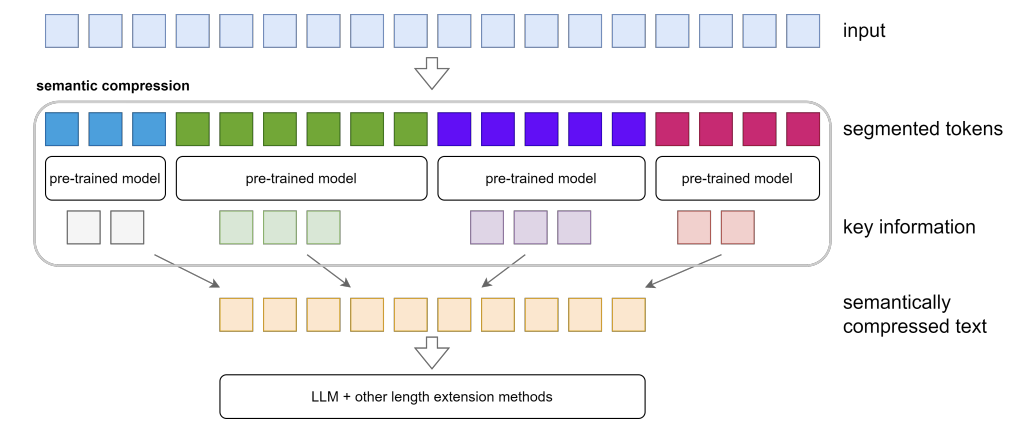}
    \caption{1) clustering the input text into thematic groups, represented as a graph, to facilitate topic-based analysis, 2) tuning the thematic segments using pre-trained models to preserve crucial details, and 3) reassembling the refined chunks in their original order - reducing the text length by approximately 6-8 times. Additionally, other techniques like extrapolation and interpolation can be used to further extend the length \cite{fei2023extending}}
    \label{fig:semantic_compression}
\end{figure}

Researchers \cite{fei2023extending} propose a semantic compression method that distills long texts into concise forms, retaining their meaning and broadening the context window (Figure~\ref{fig:semantic_compression}). This method occurs before inputting tokens into pre-trained language models and is customizable and optimized for specific tasks. It outperforms existing methods in various tasks, including question answering, summarization, and few-shot learning, without requiring additional parameter updates or memory consumption, making it computationally efficient.

\subsection{Pre-Trained Language Models (PLMs)}
The development of PLMs has revolutionized the field of NLP. The first generation of PLMs, such as Skip-Gram \cite{mikolov2013distributed}, word2vec \cite{mikolov2013efficient}, and GloVe \cite{pennington2014glove}, used shallow neural networks \cite{qiu2020pre} to obtain word embeddings. The second generation, including CoVe \cite{mccann2017learned}, ELMo \cite{peters2018dissecting}, BERT \cite{devlin2018bert}, and GPT \cite{radford2018improving}, focused on learning dynamic word embeddings using transformers. The pre-training and fine-tuning approach has achieved remarkable success in various NLP tasks. Moreover, recent breakthroughs in prompt learning \cite{liu2023pre} have empowered PLMs to accomplish few-shot or zero-shot learning with minimal labeled data. Notable examples of successful PLMs include ChatGPT, GPT-4, Gemini, Claude, LlaMA-3, Mixtral, etc.

\subsubsection{AutoCompressors} 
The authors of \cite{chevalier2023adapting} propose teaching PLMs to compress text into summary vectors \cite{lester-etal-2021-power}, which are significantly shorter than the original text (often 1-2 orders of magnitude shorter). These vectors have a two-pronged function: 1) they allow the LM to handle long documents by extending its context window with minimal computational overhead, and 2) they accelerate inference for pre-computed and cached text.

\begin{figure}[ht]
    \centering
    \includegraphics[width=\columnwidth]{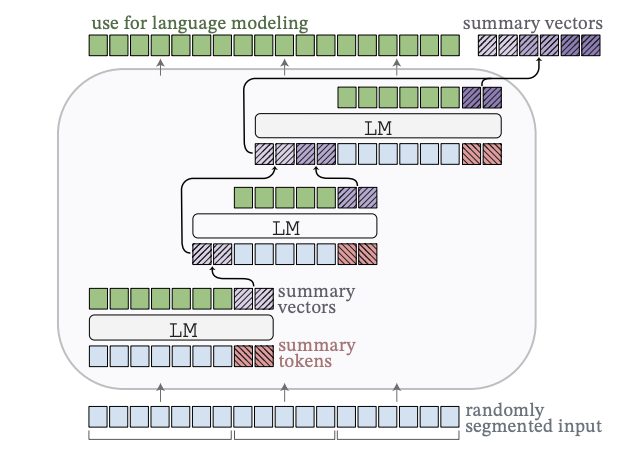}
    \caption{AutoCompressors recursively generate summary vectors from long documents, using them as soft prompts for subsequent segments \cite{chevalier2023adapting}}
    \label{fig:autoCompressors}
\end{figure}

AutoCompressors, proposed by \cite{chevalier2023adapting}, are trained To distill key information into summary vectors, generated sequentially from extended documents (Figure~\ref{fig:autoCompressors}). The approach builds upon the Recurrent Memory Transformers (RMT) architecture \cite{bulatov2022recurrent}, introducing summary accumulation and training with randomly segmented inputs. This enhances long-range information retention and facilitates reasoning across multiple passages. AutoCompressors can be seeded with PLMs and fine-tuned on long sequences. They improve perplexity for long documents and demonstrate robust compression capabilities across different domains, making them valuable for various downstream applications.

\subsubsection{LongNET}
Overcoming sequence length limitations in language models has several advantages, including improved interactions with human language, better capture of complex causality and reasoning, and reduced catastrophic forgetting. However, scaling up sequence length poses a challenge in balancing computational complexity and model expressivity. RNN-style models and state space models \cite{gu2021efficiently}, \cite{smith2022simplified}, \cite{fu2022hungry}, \cite{poli2023hyena} have been proposed, but they have limitations from the perspective of parallelization and model adaptability \cite{fathi2023block}. An alternative approach is to reduce the complexity of Transformers \cite{vaswani2017attention}, such as using sliding windows or convolution modules for attention, or sparse attention. LongNet \cite{ding2023longnet}, a novel approach, replaces the attention mechanism with "dilated attention", which achieves linear computational complexity and logarithmic dependency between tokens. This allows LongNet to efficiently scale sequence lengths to 1 billion tokens, overcoming the constraints of computation and memory.

\subsubsection{In-Context Auto-Encoders}
\begin{figure}[ht]
    \centering
    \includegraphics[width=\columnwidth]{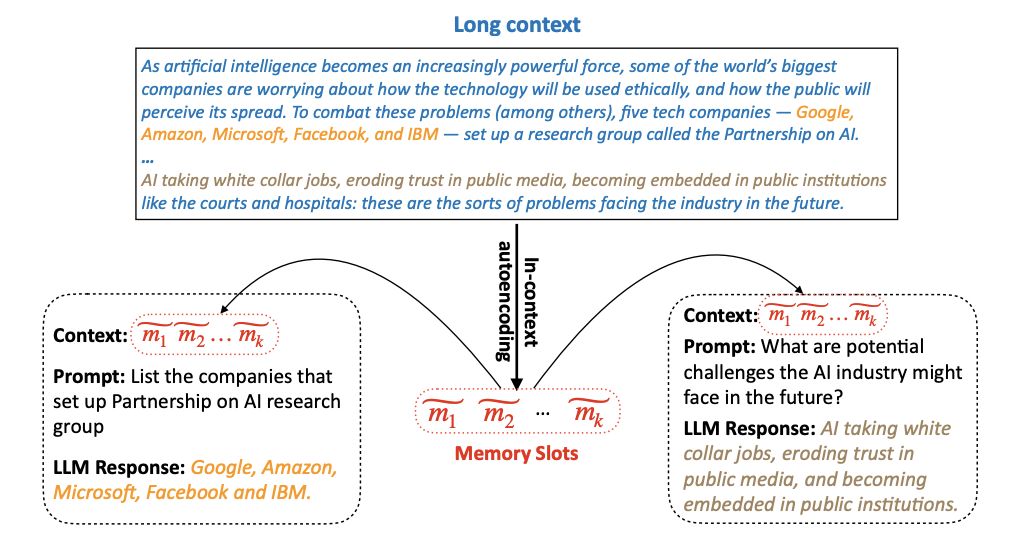}
    \caption{Condensing an extended context into a compact memory representation, which can be leveraged by the target LLM to respond to diverse prompts. \cite{ge2023context}}
    \label{fig:icae}
\end{figure}

Modeling long-range dependencies is a hurdle for Transformer-based LMs \cite{vaswani2017attention} due to their self-attention mechanism. Previous research by \cite{beltagy2020longformer}, \cite{bulatov2022recurrent}, and Ding \cite{ding2023longnet} has attempted to cope with this issue through architectural innovations, but these approaches often struggle to maintain performance in long contexts, as underscored by \cite{liu2024lost}. A novel approach, "context compression", is proposed by \cite{ge2023context}, which recognizes that an LLM can represent the same information in varying lengths. They introduce the In-context Autoencoder (ICAE), which compresses lengthy contexts into a fixed number of memory buffers using a learnable encoder and a fixed decoder (Figure~\ref{fig:icae}). The ICAE is pre-trained using auto-encoding and language modeling objectives and fine-tuned using instruction data. The approach achieves 4x context compression while maintaining effective conditioning for the target LLM, enabling faster and more memory-efficient inference.

\subsubsection{RECOMP}
\begin{figure}[ht]
    \centering
    \includegraphics[width=\columnwidth]{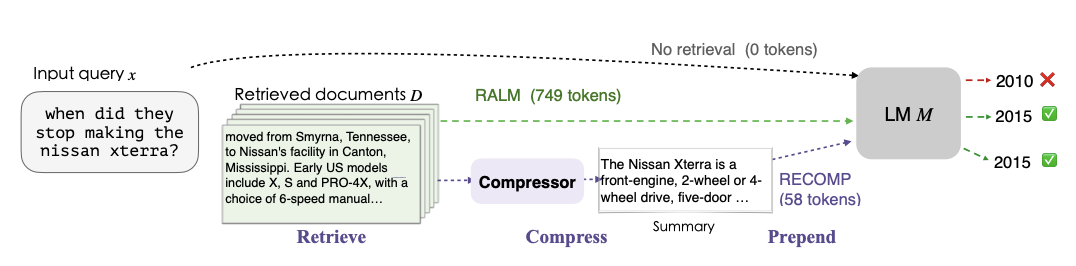}
    \caption{RECOMP's document compression technique generates a summary that serves as input to a language model, facilitating correct answer generation while minimizing encoding costs. \cite{xu2024recomp}}
    \label{fig:recomp}
\end{figure}

In their work, \cite{xu2024recomp} introduce RECOMP, an intermediary step for Retrieval-augmented Language Models (RALMs) \cite{izacard2022atlas}, \cite{borgeaud2022improving}. RECOMP compresses retrieved documents into concise textual summaries before integrating them during inference, reducing computational costs and alleviating the burden on LMs to process lengthy documents. The aim is to produce summaries that balance brevity and fidelity to the original evidence documents, guiding the RALM to produce targeted outputs when the summary is used as a prefix to the input (illustrated in Figure~\ref{fig:recomp}). To achieve this, the authors train two types of compressors:
\begin{enumerate}
\itemsep0em
    \item \textit{Extractive Compressor:} This compressor filters out irrelevant sentences, retaining only the most pertinent ones from the retrieved document set.
    \item \textit{Abstractive Compressor:} This compressor produces a summary by fusing information from multiple retrieved documents.
\end{enumerate} 
Both compressors employ a multi-document query-based summarization approach  \cite{xu2020coarse}, summarizing evidence documents concerning the input query. The authors develop training strategies that maximize performance on the target task to guarantee accurate output. Contrastive learning is employed to train the extractive compressor enabling it to select key sentences effectively, while the abstractive compressor is distilled \cite{west2021symbolic} from a large language model (like GPT-3 or GPT-4), achieving strong summarization performance. This approach holds promise for enhancing the efficiency and efficacy of RALMs.

\subsection{Retrievers}
The retriever \cite{langchain} is an interface that processes an unstructured query and returns a curated list of documents in response. Contextual compression aims to address the challenges of retrieval by compressing the retrieved context to only include relevant information. In this context, "compressing" encompasses both condensing the content of individual documents and eliminating irrelevant documents altogether. The Contextual Compression Retriever uses a \textit{base retriever} and a \textit{Document Compressor} to process queries. The base retriever retrieves the initial documents, which are then passed through the Document Compressor to shorten the list of documents by either reducing the contents of individual documents or excluding entire documents altogether.

\subsubsection{LLMChainExtractor}
In this approach, the base retriever is wrapped with a \textit{ContextualCompressionRetriever}. Additionally, an \textit{LLMChainExtractor} serves as the base compressor. The \textit{LLMChainExtractor} iterates over the initially retrieved documents and extracts only the relevant content for the given query. It achieves this by making an additional LLM call for each retrieved document and summarizing the relevant information

\subsubsection{EmbeddingsFilter}
Making an additional LLM call for each retrieved document can be both costly and slow. However, the \textit{EmbeddingsFilter} offers a more economical and faster alternative. By embedding both the documents and the query, it selectively returns only those documents that exhibit sufficiently similar embeddings to the query. This approach optimizes retrieval efficiency while maintaining relevance.

\subsubsection{DocumentCompressorPipeline}
The DocumentCompressorPipeline allows a seamless combination of multiple compressors in a sequence. Alongside these compressors, we can incorporate \textit{BaseDocumentTransformers} into our pipeline. Unlike contextual compressors, these transformers don’t alter the content significantly but perform specific transformations on a set of documents. For instance, \textit{TextSplitters} can divide documents into smaller segments, while the \textit{EmbeddingsRedundantFilter} identifies and filters out redundant documents based on embedding similarity. This modular approach enhances flexibility and adaptability in document processing. e.g.
\begin{itemize}	
\itemsep0em
    \item \textit{Splitter:} create small chunks
    \item \textit{Redundant filter:} remove similar docs — embedded
    \item \textit{Relevant filter:} relevant to query
\end{itemize}

\section{Metrics and Benchmarks}

\subsection{Metrics}
\begin{figure}[ht]
    \centering
    \includegraphics[width=\columnwidth]{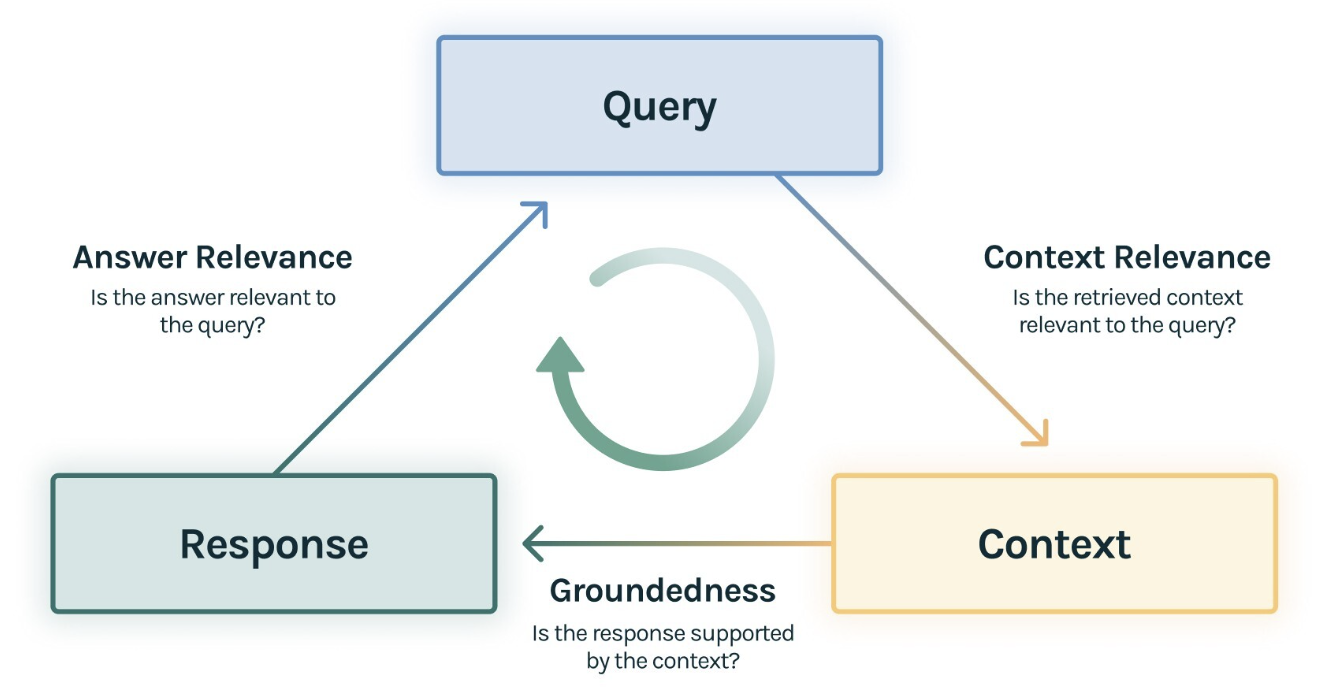}
    \caption{RAG-Triad}
    \label{fig:rag_triad}
\end{figure}

Evaluating language model inference efficiency involves considering various metrics that capture different performance aspects, including accuracy, zero-shot capabilities, compression ratio, and inference time. Within the framework of RAG-based solutions, the "Triad of Metrics" \footnote{RAG Triad (Figure~\ref{fig:rag_triad}): \url{https://www.trulens.org/trulens_eval/getting_started/core_concepts/rag_triad/}} - Groundedness, Context Relevance, and Answer Relevance - are also employed for evaluation. Achieving satisfactory performance across these metrics helps ensure that the language model application is reliable and free from hallucinations.

\subsubsection{Compression Ratio}
The compression ratio measures the reduction in size from the original uncompressed context to the compressed context. A higher compression ratio means that the compression is more efficient, as it achieves a greater reduction in size while preserving the context's coherence.

\subsubsection{Inference Time}
Inference time, also known as latency, measures how long it takes for a Large Language Model (LLM) to process input data and generate responses. This metric is crucial for real-world applications that require quick handling of user queries or processing of large data volumes in real-time.

\subsubsection{Context Relevance}
In RAG applications, the first step is retrieval, and it's crucial to ensure that the retrieved context chunks are relevant to the input query. Irrelevant information in the context can lead to hallucinations in the LLM's answer. To evaluate context relevance, the structure of the serialized record can be analyzed.

\subsubsection{Groundedness}
After retrieving the context, an LLM transforms it into an answer. However, LLMs can sometimes stray from the facts and generate responses that are not entirely accurate. To ensure the groundedness of the application, the response can be broken down into individual claims and verified by searching for supporting evidence within the retrieved context.

\subsubsection{Answer Relevance}
Furthermore, our response must still effectively address the original question. We can assess this by evaluating the relevance of the final response to the user’s input.

\subsubsection{Others}
RAG evaluation also encompasses four key abilities that reflect the model’s adaptability and efficiency: noise robustness, negative rejection, information integration, and counterfactual robustness \cite{chen2024benchmarking}, \cite{liu2023recall}. The model's quality scores are heavily influenced by its ability to leverage these capabilities in diverse challenges and complex scenarios:
\begin{enumerate}
\itemsep0em
    \item \textit{Noise Robustness:} This metric gauges a model's capacity to distinguish between relevant and irrelevant documents, even when the latter are tangentially related to the question.
    \item \textit{Negative Rejection:} The metric measures a model's capacity to recognize when the retrieved documents are insufficient to answer a question, and to withhold a response accordingly.
    \item \textit{Information Integration:} Information integration tests a model's proficiency in combining relevant information from multiple documents to provide well-informed answers to challenging questions.
    \item \textit{Counterfactual Robustness:} Counterfactual robustness measures a model's skill in identifying and ignoring flawed or misleading information in documents, regardless of its awareness of potential errors.
\end{enumerate} 
In brief, context relevance and noise robustness are crucial for evaluating the retrieval process, while answer groundedness, answer relevance, negative rejection, information integration, and counterfactual robustness are vital for assessing the quality of generated text.

\subsection{Benchmarks and Datasets}
The primary objective of these benchmarks and datasets is to assess the trade-offs between compressed and uncompressed contexts in terms of effectiveness, efficiency, and accuracy, covering a broad range of NLP tasks and applications.

\subsubsection{Common Benchmarks and Datasets}
RAG's primary function revolves around answering questions, encompassing various formats such as single-hop and multi-hop queries, multiple-choice options, and domain-specific inquiries, as well as lengthy scenarios that leverage RAG's capabilities. Moreover, RAG is constantly evolving to tackle additional tasks, including extracting relevant information, generating conversational dialogue, and searching for code snippets, documentations and even interpreting them. For more details, refer to the study by \cite{gao2023retrieval}.

\section{Challenges and Future Directions}
\subsection{More advanced Methods}
Research on contextual compression for LLMs is still in its early stages. While previous studies have shown compressed contexts, they still lag behind uncompressed contexts in terms of performance. By exploring more advanced compression methods tailored for LLMs, we can potentially bridge this performance gap and enhance the performance of uncompressed contexts.

\subsection{Performance-Size Trade-offs}
Previous research highlights the importance of balancing LLM performance with context size, considering hardware limitations and practical constraints. Despite its significance, the theoretical and empirical foundations of this trade-off remain poorly understood. Future investigations should focus on conducting exhaustive examinations to drive the creation of sophisticated compression techniques that can meet the demands of increasingly complex data sets, enabling researchers to create tailored methods that effectively navigate the design space and optimize performance.

\subsection{Dynamic Contextual Compression}
Contemporary compression approaches still utilize manual compressors, such as retrievers, which often require an empirical methodology driven by input data or task specifications. This can be a practical hindrance to adoption, especially in scenarios like context distillation, where finding suitable student templates within computational constraints can be time-consuming and require multiple trials.

\subsection{Explainability}
Compressing pre-trained language models can make them hard to understand (lacking explainability). To fix this, using explainable compression methods can help make models more interpretable, easier to evaluate, and more reliable in real-life scenarios.

\section{Conclusion}
This in-depth analysis explores the domain of contextual compression techniques, with a focus on their application to LLMs. Our study encompasses a broad range of compression methods, evaluation metrics, and benchmark datasets, providing a comprehensive understanding of the field. By examining the complexities of contextual compression, we identify the key challenges and opportunities that arise in this area. As research in this field continues to advance, the development of specialized methodologies tailored to the needs of LLMs is crucial for unlocking their full potential across various domains. This survey aims to serve as a valuable resource, providing a detailed overview of the current landscape and encouraging further investigation into this vital topic.

\section*{Limitations}
While this survey provides a comprehensive overview of contextual compression techniques for large language models, there are several limitations to acknowledge. Firstly, the field of contextual compression is rapidly evolving, and this survey may not capture the very latest advancements in the area. Additionally, the focus on large language models may not be representative of other types of language models or AI systems, which may have different compression requirements. Furthermore, the survey's reliance on existing evaluation metrics and benchmark datasets may not fully capture the complexities and nuances of contextual compression. Moreover, the need for advanced methodologies specifically designed for LLMs highlights the potential limitations of current approaches, which may not be scalable or effective for future LLM architectures. Finally, the survey's scope is limited to contextual compression, and future research may uncover new challenges and opportunities at the intersection of compression and other aspects of LLMs.

\section*{Ethics}
As research in contextual compression for large language models continues to advance, it is essential to consider the ethical implications of these developments. One key concern is the potential for biased or unfair compression methods, which could perpetuate existing social inequalities or create new ones. For instance, compression techniques that prioritize certain types of data or language styles over others may disadvantage certain groups or communities. Furthermore, the focus on large language models may exacerbate existing power imbalances, where only those with access to significant computational resources and data can develop and deploy these models.

Additionally, the reliance on existing evaluation metrics and benchmark datasets may perpetuate biases and limitations in the development of compression techniques. It is crucial to ensure that these metrics and datasets are diverse, representative, and regularly updated to reflect the complexities of real-world language use.

The need for advanced methodologies specifically designed for LLMs also raises ethical concerns around the responsible development and deployment of these models. As LLMs become increasingly ubiquitous, it is essential to consider their potential impact on individuals, communities, and society as a whole. This includes ensuring that these models are transparent, explainable, and accountable, and that their development and deployment are guided by ethical principles and values.

Finally, the survey's limited scope to contextual compression highlights the need for a more comprehensive consideration of the ethical implications of LLMs and their applications. Future research should prioritize ethical considerations and ensure that the development of compression techniques and LLMs is guided by a commitment to social responsibility, fairness, and transparency.

\bibliography{latex/main}

% \appendix

% \section{Example Appendix}
% \label{sec:appendix}

% This is an appendix.

\end{document}